# Patchwork Learning: A Paradigm Towards Integrative Analysis across Diverse Biomedical Data Sources

Suraj Rajendran[1], Weishen Pan[2], Mert R. Sabuncu[3,4,5], Yong Chen[6], Jiayu Zhou[7], Fei Wang[2*]

[1]Tri-Institutional Computational Biology & Medicine Program, Cornell University, NY, USA
[2]Division of Health Informatics, Department of Population Health Sciences, Weill Cornell Medicine, New York, NY, USA
[3]School of Electrical and Computer Engineering, Cornell University, Ithaca, NY, USA
[4]Cornell Tech, New York, NY, USA
[5]Department of Radiology, Weill Cornell Medical School, New York, NY, USA
[6]Department of Biostatistics, Epidemiology and Informatics, University of Pennsylvania, Philadelphia, PA, USA
[7]Department of Computer Science and Engineering, Michigan State University, East Lansing, Michigan, USA

*Corresponding author
E-mail: few2001@med.cornell.edu

**Abstract**
Machine learning (ML) in healthcare presents numerous opportunities for enhancing patient care, population health, and healthcare providers' workflows. However, the real-world clinical and cost benefits remain limited due to challenges in data privacy, heterogeneous data sources, and the inability to fully leverage multiple data modalities. In this perspective paper, we introduce "patchwork learning" (PL), a novel paradigm that addresses these limitations by integrating information from disparate datasets composed of different data modalities (e.g., clinical free-text, medical images, omics) and distributed across separate and secure sites. PL allows the simultaneous utilization of complementary data sources while preserving data privacy, enabling the development of more holistic and generalizable ML models. We present the concept of patchwork learning and its current implementations in healthcare, exploring the potential opportunities and applicable data sources for addressing various healthcare challenges. PL leverages bridging modalities or overlapping feature spaces across sites to facilitate information sharing and impute missing data, thereby addressing related prediction tasks. We discuss the challenges associated with PL, many of which are shared by federated and multimodal learning, and provide recommendations for future research in this field. By offering a more comprehensive approach to healthcare data integration, patchwork learning has the potential to revolutionize the clinical applicability of ML models. This paradigm promises to strike a balance between personalization and generalizability, ultimately enhancing patient experiences, improving population health, and optimizing healthcare providers' workflows.

**Introduction**
Machine learning (ML) in healthcare is a rapidly evolving field, presenting numerous opportunities for progress. Active and passive patient data collection, both during and outside medical care, can be utilized to address health challenges. As a result, ML has become an essential tool for processing and analyzing these data in various domains, including natural language processing, computer vision, and more. ML systems have demonstrated their potential to enhance patient experiences, improve population health, reduce per capita healthcare costs, and optimize healthcare providers' workflows[1–3]. However, the real-world clinical and cost benefits of ML in healthcare remain limited, indicating a significant gap in its application.

Data privacy is a major challenge facing the use of ML in healthcare, as it restricts the potential for pooling electronic health record (EHR) data from multiple sites. Federated learning (FL) offers a promising approach to addressing this issue by enabling the aggregation of fragmented, sensitive data from various sites without sharing any sensitive information[4–7]. In brief, a typical FL architecture consists of a central aggregator designed to obtain global ML model parameters by iteratively exchanging local ML model parameters. Despite its potential, implementing FL in healthcare comes with its own set of challenges, and while mitigation strategies have been proposed, they are not without their limitations[8,9]. Another major concern in healthcare ML systems is their inability to generalize well beyond the sites from which the training data were sourced, mainly due to heterogeneity across sample populations[10,11]. Heterogeneity can stem from various factors, such as distinct demographics or regions, differing medical workflows, temporal drifts, and variations in administrative practices[12,13]. Even when a model is trained using heterogeneous data from multiple sites, achieving generalization can still be challenging, as essential site-specific information may be required for optimal local performance. Consequently, a balance between personalization and generalizability must be struck when developing ML models for healthcare applications[14–16].

Another factor limiting the clinical applicability of ML models is their inability to fully utilize all available data modalities. While single modality models exist (e.g., clinical notes, lab tests, omics, or medical images), systems that simultaneously leverage multiple modalities are relatively scarce. Furthermore, these multimodal models often lack strategies for clinical deployment and bias mitigation[17]. Ideally, these systems should integrate complementary information from various modalities, providing a more comprehensive representation of data for ML algorithms to complete tasks more effectively. By offering a more holistic perspective on available data, these models can better mimic human interactions with data, considering multiple aspects of a problem simultaneously[18–21]. To facilitate the development of such systems, a paradigm called multimodal/multiview learning (MML) has been proposed. MML combines disparate data sources to capitalize on complementary information, thereby improving performance. Although MML has recently gained attention in healthcare, several issues have limited its adoption, including the difficulties in integrating multiple data sources with diverse formats, the need for large annotated datasets, and the challenge of meaningfully combining diverse information sources for enhanced performance. Moreover, many current MML models in healthcare integrate information from only two or three modalities, failing to take advantage of the vast array of medical modalities available.

In this perspective, we introduce a paradigm we term "patchwork learning" (PL). We explore the concept of patchwork learning and its current implementations in healthcare. Subsequently, we discuss numerous opportunities where patchwork learning can be applied within healthcare, along with potential data sources for addressing these areas. We also examine the challenges associated with patchwork learning, many of which are shared by federated and multimodal learning. Finally, we conclude with recommendations for future research in this field, specifically in the context of healthcare.

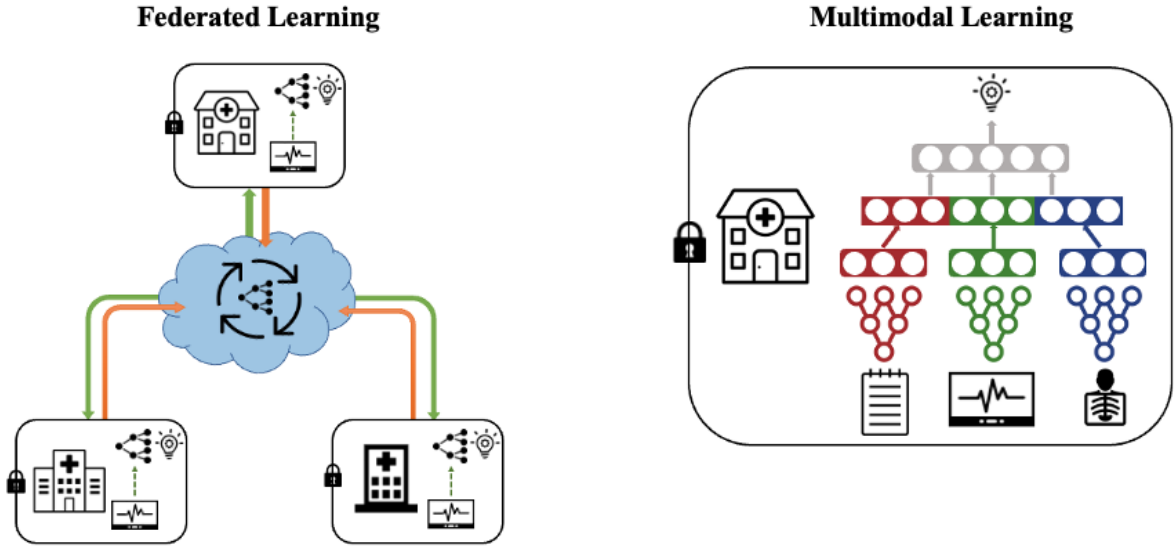

*Figure 1: Federated and multimodal learning diagrams*

**Patchwork Learning**

Patchwork learning is the concept of integrating information from disparate datasets composed of different data modalities (e.g., clinical free-text, medical images, omics) and distributed across separate and secure sites (i.e., data from Silo 1 has a distinct set of patients compared to Silo 2). Some sites may not possess all data modalities required for model development, resulting in a patched setting (Figure 2). While certain sites may lack specific data modalities, there must be some level of overlap between feature or sample spaces across sites, i.e., bridging modalities and/or patients that link different sites together. In scenarios where these bridges are unavailable, external information may be necessary to create them.

More formally, consider a patchwork involving $N$ sites with $K$ modalities, where different sites may have varying subsets of available modalities. Let $M = \{M_1, ..., M_K\}$ represent the set of all possible modalities across all sites. Note that not all modalities are accessible at each site. Alongside the multi-modality data, multiple related prediction tasks exist across all sites. The goal of patchwork learning is to impute the missing modalities and subsequently address these prediction tasks.

The relationships between modalities in patchwork learning can be depicted using a graph. Assume each modality $M_p$ is represented by a node in the graph, with an undirected edge connecting $M_p$ to $M_q$ if at least one client possesses data from both modalities. An edge between $M_p$ to $M_q$ signifies the possibility of modeling the bridge between them. To facilitate the imputation of all missing grids in a patchwork, we assume that all observed modalities in the patchwork form a connected graph. A connected graph will allow the information to be shared across sites and modalities. However, it is important to acknowledge that this assumption may not always hold.

Integrating diverse healthcare datasets poses a significant challenge, as missing data often compromises the quality and accuracy of machine learning models. Within a single site or domain, non-random missingness patterns correlated with outcomes may emerge, leading to some samples lacking specific feature types. Numerous state-of-the-art solutions have been proposed to address this issue, including imputation methods utilizing statistical techniques (e.g., nearest-neighbor interpolation, soft imputation, and multivariate imputation by chained equations) and data augmentation techniques that generate synthetic data to supplement incomplete datasets[22,23]. However, in the context of PL, addressing missingness within a single site or domain is insufficient, as entire data modalities may be absent at certain sites. PL seeks to integrate information from multiple sites, each characterized by its distinct patient population and data modalities. In this situation, missingness extends beyond individual samples to encompass missing modalities across sites. The challenge entails securely modeling the relationships between different modalities, inferring, and integrating them to construct models that generalize across multiple sites without sharing raw data. While traditional methods for addressing missingness within a single site or domain can be helpful, they fall short in overcoming the complex challenges presented by PL. Patchwork learning necessitates a more sophisticated approach, adept at effectively managing missing modalities across various sites and integrating disparate datasets to create robust machine learning models.

Adopting PL frameworks has the potential to advance healthcare in several ways. As previously mentioned, models developed at one site are not readily applicable to another site due to data heterogeneity. Incorporating various data modalities across different sites can mitigate some biases that currently affect ML models. Furthermore, in real-world scenarios, certain organizations may not have access to multiple modalities. Secure information sharing across sites can facilitate the development of robust models capable of integrating multiple modalities of healthcare data, even at organizations lacking those modalities. A patchwork learning formulation can prove helpful in situations where different sites have related but distinct tasks, as it may be possible to train a shared model backbone for downstream tasks. The backbone model can be pre-trained through a patchwork learning framework using all available modalities and patients across different sites. The pre-training process can be implemented using a general-purpose approach, such as employing a self-supervised strategy, which facilitates the learning of informative representations applicable to a wide range of downstream tasks. After training the backbone model, it can be fine-tuned to cater to specific tasks at each site.

When formulating a patchwork learning problem, several key considerations must be taken into account. These include addressing differences in data distribution across sites, inferring data modalities that are absent at specific sites, and effectively integrating these modalities to develop robust models. We highlight a few practices in the literature that we deem suitable for patchwork learning, which address some of the aforementioned challenges.

Federated transfer learning (FTL) is a specialized form of FL that employs distinct datasets, differing not only in sample space but also in feature space. FTL allows users to address data distribution differences across clients effectively[24,25]. A core component of FTL is transfer learning, a machine learning technique that aims to enhance the performance of target models developed on target domains by reusing the knowledge contained in diverse but related models developed on source domains. FTL can be performed in multiple ways, but generally, knowledge across sites is securely transferred and/or aggregated, despite differing feature spaces between sites. Overall, there are two strategies in FTL: 1) using pre-trained models in related tasks, or 2) using domain adaptation to transfer knowledge from a source domain to a related target domain. Chen et al. developed an FTL algorithm, FedHealth, which uses domain adaptation to analyze multimodal healthcare data from wearables. To address the data isolation and heterogeneity issues associated with wearable data, FedHealth first trains a model on public data at the central server, which it then transfers to clients iteratively for further personalization[26]. To apply federated transfer learning to patchwork learning, the relationships between modalities can be regarded as the knowledge to be transferred across sites. Following FedHealth's methodology, models capable of inferring missing modalities can be trained on public data and adapted to each site. A crucial aspect of this approach is ensuring that the public data encompasses a comprehensive range of potential modalities across all sites.

As previously discussed, clients in a PL setting are likely to lack all data modalities at their site, which constrains their ability to develop integrated models. Confederated learning provides a solution to this issue[27]. In confederated learning, machine learning models are trained on data distributed across diverse populations and data types, employing a three-step approach. The concept was introduced in a study where a patient population's data was split both horizontally and vertically, i.e., different sites had varying combinations of data (diagnostic data, medications, lab tests) and patients. Notably, confederated learning requires an auxiliary dataset to be available at the central server, which may not be realistic in real-world scenarios. To perform confederated learning, conditional generative adversarial networks with matching loss (cGAN) were trained using data from the central server to infer one data type from another. These cGANs are transferred to each local site where the missing data types are imputed with generated samples. Thereafter, task-specific models, like diagnosis prediction, were trained in a federated manner (e.g., federated averaging) across all sites simultaneously. Compared to other methods, confederated learning is simple to implement and does not require any patient ID matching. A core weakness, however, is that the success of the approach depends on the quantity and heterogeneity of data available at the central server. The performance of the subsequent confederated model can be affected by any discrepancies between the auxiliary data and the data at each local site[27]. One potential method for applying federated learning to patchwork learning involves utilizing a variant of the cGAN, known as the Conditional Wasserstein Generative Adversarial Network with Gradient Penalty (CWGAN-GP). The CWGAN-GP has demonstrated effectiveness in generating realistic samples even when faced with limited training data, a common scenario with auxiliary data stored at a central server in healthcare settings. In this approach, the CWGAN-GP generates missing data modalities for each client, provided that the shared modalities at the central server cover all K modalities across clients. The synthesized modalities can subsequently be employed to train a downstream model tailored for that client, such as a classifier or regressor.

In a PL setting, even when all modalities are available, it is essential to effectively integrate these modalities for modeling downstream tasks. Several studies have explicitly combined FL and MML to develop federated multimodal learning (FML) systems[28]. These systems are designed to integrate multiple data modalities across sites. Some personalized recommendation systems use FML with multimodal matrix factorization methods to provide privacy-preserving predictions based on both text and demographic data[29,30]. Salehi et al. propose FLASH, which fuses data from LIDAR, GPS, and camera images to train a federated model across vehicles, optimizing vehicular communication transmissions[31]. Another FML method, FedMMTS, uses multimodal analytics to create privacy-preserving systems that enable autonomous decision-making for vehicles in a simulated environment[32]. While FML systems have been applied in multiple domains, their implementation in healthcare has been limited. Challenges associated with missing data, patient privacy, and the need for clinical interpretability constrain the adoption of FML in healthcare. Che et al. design H-FedMV and S-FedMV, which perform FML across sites using federated averaging, the latter of which is able to account for sequential information within medical data. Modalities that were integrated include textual and time-series data, both of which were used to diagnose patients with bipolar disorder[33]. Another study utilized an FML system to predict oxygen requirements for COVID-19 symptomatic patients by combining data from EHRs and chest X-rays. To integrate the different modalities and increase the interaction between data types, a Deep & Cross network architecture was used across all sites, followed by fully connected layers for performing prediction. To add privacy-preserving measures to their model, differential privacy was implemented in the federated weight-sharing mechanisms. While the model performed relatively well on validation data, the architecture required the presence of all modalities at all sites, which is unrealistic in a real-world scenario without losing a considerable amount of available data[20]. In a patchwork learning setting, FML systems can be designed to integrate multiple data modalities across sites, by using federated averaging, or more complex methods, to train a shared model backbone on the available modalities at each site within a fully connected graph. Once the shared model backbone is trained, it can be fine-tuned for specific tasks at each client site, by training a task-specific model on the available modalities at each site, along with generated missing modalities.

The aforementioned methods can be suitable for the patchwork learning paradigm, where the goal is to learn from multiple data modalities distributed across secure silos. Each method possesses distinct weaknesses and strengths, which we will examine in the following section. However, these approaches collectively demonstrate potential for advancing the development of robust and generalizable machine learning models in the healthcare domain.

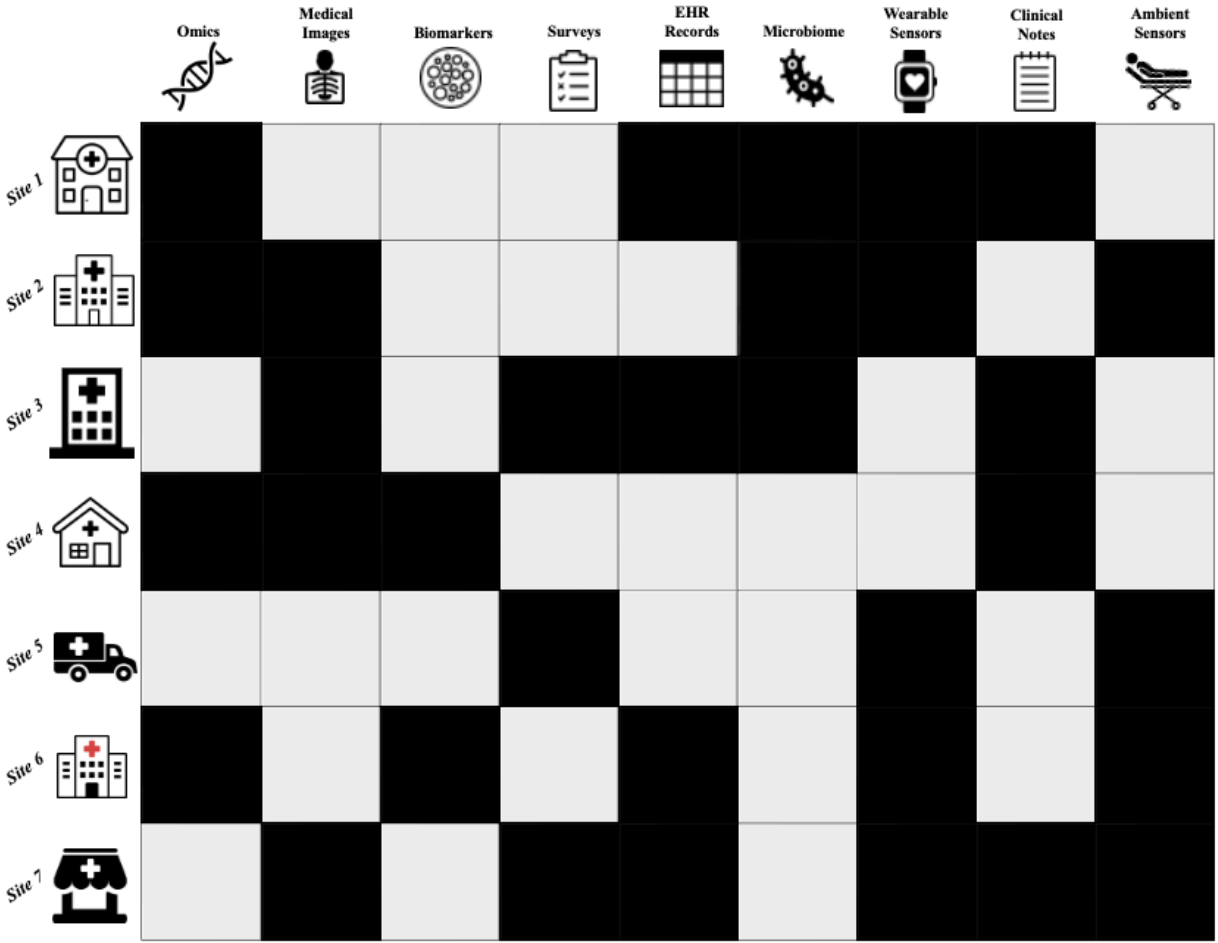

*Figure 2*: Overview of patchwork learning in healthcare

## Opportunities for Patchwork Learning

**Predictive Diagnosis and Risk Prediction**
Predictive analytics has emerged as a valuable tool in medical decision-making, as patients' responses to treatments, particularly for chronic diseases, can vary significantly. Machine learning algorithms, using data and outcomes from past patients, can offer insights into the most effective treatment methods for current patients. Most cutting-edge clinical risk predictive models are based on deep learning and trained end-to-end. However, the robustness of diagnostic or risk prediction tools relies heavily on the breadth of data used to build them. Diagnostic ML models developed using site-specific unimodal data often face challenges when adapting to other clinics[18,19,21]. This issue persists even when incorporating federated methods due to heterogeneity and lack of local personalization[34,35].

The integration of multiple modalities has improved the performance of these algorithms across different sites. For example, stroke manifestations can be found in both EHR and medical imaging data, indicating that combining both could result in more accurate risk prediction models. Boehm et al. used a multimodal dataset, including CT scans, H&E-stained pathology slides, omics, and clinical data, to stratify risk for patients with high-grade ovarian cancer. This approach revealed the complementary prognostic information provided by each modality[36]. Ali et al. combined data from sensors and EHRs to detect cardiovascular disease while generating automated recommendations for patient care[37]. However, medical data is often siloed, making it difficult to access large multimodal datasets. In such situations, PL can facilitate the development of robust models trained on heterogeneous and multimodal datasets. Qayyum et al. introduced Clustered Federated Learning (CFL), an algorithm that uses a federated multi-tasking framework to group clients into modality-dependent clusters with jointly trainable data distributions for COVID-19 diagnosis prediction. CFL outperformed other unimodal federated models[38]. Another study developed a generalized federated multimodal disease diagnosis prediction model using a fusion and least-squares algorithm, which significantly outperformed locally trained unimodal counterparts[39]. Agbley et al. employed a MMFL framework to create a melanoma detection model using matched EHR data and skin-lesion images[40]. Alam et al. developed FedSepsis, a model for early detection of sepsis that incorporated tabular and textual data from EHRs using federated multimodal learning. The study used low-computational edge devices, such as Raspberry Pi and Jetson Nano, to address practical challenges[41].

Models developed in PL-like settings have demonstrated comparative advantages over both unimodal federated models and multimodal single-institution models for risk prediction. While heterogeneity across silos still needs to be addressed when developing PL-based models, these studies have shown clinical potential in terms of performance and interpretability. Similar to physicians' decision-making processes, PL enables models to learn risk factors from multiple modalities while accounting for hospital-dependent confounders.

**Personalized Omics for Precision Health**
The advent of next-generation sequencing technologies has led to increased interest in studying human health by interpreting molecular intricacies and variations at multiple levels, such as the genome, epigenome, proteome, and metabolome. This omics data integrates large amounts of personalized patient data, which is crucial for understanding individual disease states, distinguishing sub-phenotypes, and developing digital twins, among other applications[42–44]. Machine learning methods offer innovative techniques for integrating various omics data to discover new biomarkers, potentially aiding in accurate disease prediction and precision medicine delivery. The study of integrative machine learning methods for multi-omics data enables a deeper understanding of biological systems during normal physiological functioning and in the presence of disease, supporting insights and recommendations for interdisciplinary professionals. Shen et al. recently presented a strategy for capturing and analyzing multiple molecular data types from just 10 μl of blood, including thousands of metabolites, lipids, cytokines, and proteins. This approach was enhanced by integrating physiological information from wearable sensors[45]. However, the study is limited by its small dataset size, which may not be representative of the general population and may lack statistical power. Patchwork learning could address these limitations by collecting data from various sites and integrating it, allowing for a larger, more representative sample size.

Automated pipelines like GenoML, developed by Makarious et al., enable users to analyze multi-omics data in combination with clinical data while providing a federated module for basic privacy-preserving omics analysis across data silos[46]. Many current multimodal methods incorporating omics data use matrix factorization methods or canonical correlation analysis (CCA) to combine information from multiple modalities[39]. LungDWM uses multi-omics data to diagnose lung cancer subtypes by fusing omics-specific features extracted from an attention-based encoder. Missing omics-specific features are imputed through generative adversarial learning[47].

Despite the wealth of data available, several challenges hinder the development of omics-based models, including heterogeneity across omics technologies, missing values, interpreting multilayered system models, and issues related to data annotation, storage, and computational resources. Additionally, omics data is inherently heterogeneous and high-dimensional, leading to limited statistical power in many investigations due to small dataset sizes[39]. Patchwork learning formulations can help address these challenges by integrating various omics data along with other medical patient information across multiple secure silos. Zhou et al. propose PPML-omics, a federated learning pipeline for integrating and analyzing omics data using differential privacy for added security[48].

Currently, no platform can efficiently integrate clinical, multi-omics, and other data modalities while simultaneously enabling effective management of data analytics accessible to physicians[49,50]. Successful integration of different omics data with other data types, such as EHRs and medical images, has the potential to enhance our understanding of a patient's health, allowing for the development of personalized preventative and therapeutic interventions. Such integrations require big data platforms or methodologies that facilitate fusion of heterogeneous modalities from multiple silos while allowing real-time care[51]. Developing these systems requires not only technologies following patchwork learning paradigms but also new regulations and business models that promote their use in precision medicine[52].

**Digital Clinical Trials**

The ever-increasing sources of clinical data from EHRs, claims, and billing data have generated massive amounts of real-world data (RWD) with the potential for translational impacts on patients. In recent years, trial emulation, the process of mimicking targeted randomized controlled trials (RCT) with real-world data such as electronic health records, has gained attention in the medical community. While RWD is more representative of real patient populations, numerous challenges are associated with conducting trial emulation, such as identifying and controlling confounding variables, constructing proper RCT designs, and determining appropriate causal inference methods for outcome estimation. Although there is a growing body of research addressing these challenges, data access remains a significant limitation, especially for trial emulations focused on less common conditions and treatments. Gaining access to RWD can be a lengthy and costly process, and due to privacy concerns, aggregating private health data, which is often richer in information on specific conditions, can be difficult. Accessing RWD from various clinical sites can help combat data heterogeneity in patient populations, allowing trial emulation hypotheses to be generalized across demographic and geographic groups[53]. Liu et al. introduced the Distributed Algorithm for fitting Penalized (ADAP) regression models to integrate patient-level data from multiple sites, studying risk factors for opioid use disorder. To securely share information and mitigate heterogeneity across multiple sites, collaborating sites only share first and second-order gradients when conducting trial emulation[54].

Integrating data from wearable technologies can also improve trial emulation outcomes. Readings from wearables, such as sleep, physical activity, vital signs, and questionnaires, can provide valuable information for balancing confounders during trial emulation pipelines, despite being noisy. Machine learning techniques can be employed to integrate data from wearables, omics, EHRs, and medical images for digital clinical trials[55–57]. However, the lack of interoperability between real-world databases currently limits the performance of multimodal trial emulation pipelines, and architectures capable of simultaneously leveraging longitudinal RWD from multiple modalities are not yet available[58]. Despite the high cost of conducting clinical trials, causal inference with patchwork learning can help identify pertinent medications or treatments through trial emulation. By integrating heterogeneous sources of data, both in terms of features and samples, confounding variables can be controlled, enhancing the capabilities of digital clinical trials. SurvMaximin is one algorithm in this field that combines multiple prediction models from different source outcomes in a federated manner for predicting survival outcomes. Importantly, different clients can have non-uniform feature spaces, enabling a patchwork learning paradigm for determining survival outcomes[59].

Recent investigations have delved into developing digital twins, virtual representations of physical entities, such as medical patients, that can be used to simulate and analyze real-world scenarios[43]. Within the framework of patchwork learning, a digital twin can be created using various data sources, including medical records, imaging scans, and genetic information. By combining heterogeneous data across sites, digital twins can offer highly detailed and accurate representations of patients. Using trial emulation with digital twins, doctors and researchers can test different treatment options and predict individual patient outcomes, allowing for more personalized and effective care. Digital twins can also serve as synthetic control arms in randomized clinical trials and trial emulation studies. In a randomized clinical trial, the control group typically receives a placebo or standard treatment, while the experimental group receives the treatment being tested. However, in certain cases, using a placebo or standard treatment as a control may not be ethical or feasible, particularly when targeting high-risk patients. In these situations, a digital twin, generated using patchwork learning, can act as a synthetic control arm. The digital twin can simulate the control group and compare outcomes with the experimental group, leading to more accurate and ethical clinical trials and trial emulation studies. Additionally, digital twins can be employed to monitor patient progress over time and identify potential issues before they arise, improving overall patient outcomes.

As the costs of real-world clinical trials continue to rise, computational tools will be essential for supplementing hypothesis generation. Confounding patient and environmental variables, spread across multiple data modalities, must be accounted for even in extensive collections of RWD. Patchwork learning formulations can offer unique ways to mitigate confounding variables and integrate private data sources, enabling hypothesis generation for rare medical conditions.

**Remote Monitoring**
Medical Internet of Things (MIoT) devices, such as wearable technologies and mobile phones, enable continuous monitoring of human physical activities, behaviors, and physiological and biochemical parameters during daily life. Commonly measured data include vital signs like heart rate, blood pressure, and body temperature, blood oxygen saturation, posture, and physical activities, gathered through electrocardiogram (ECG), ballistocardiogram (BCG), and other devices. Monitoring physical activity and interactions has numerous use cases in preventing adverse health outcomes. Recently, data from wearables have been used to predict COVID-19 symptoms remotely[60]. Algorithms for monitoring mental conditions using wearables have also gained attention. Some wearable devices are equipped with sensors that detect human physiology status, such as heartbeat, blood pressure, body temperature, or other complex vital signs (e.g., electrocardiograms). Using these signals, new systems can be developed to monitor mental conditions. Xu et al. developed FedMood, which fuses keystroke and accelerometer data from mobile phones to create a federated model for medical depression detection[61]. Fed-ReMECS combines electrodermal, respiration, and EEG signals to perform real-time emotion state classification across several secure wearable sensors. Their architecture is scalable to include other data modalities crucial for emotion detection[62]. Liang et al. present a privacy-preserving multimodal model trained on typed text, keystrokes, and app usage from a patient's phone. By combining different modality features through late fusion and using a selective-additive-learning framework for privacy preservation, their model performs mood assessments better than unimodal models[63]. Wearable technology can also improve patient management efficiency in hospitals by providing early detection of health imbalances. Wireless communication in wearable techniques enables researchers to design a new breed of point-of-care (POC) diagnostic devices[64–66].

However, processing information from wearables can be challenging. Data from wearables is intrinsically multimodal, ranging from audio and images to time-series data. Although efforts have been made to fuse data types from various sensors, there are further advantages to combining wearable data with formalized clinical data, like those recorded in EHRs. Wang et al. propose an architecture design for COVID-19 diagnosis using a combination of demographic information, medical record text data, patient mobile data, and image data stored across different nodes. While not implementing this design, they highlight the advantages of enabling such architectures for real-time pandemic monitoring[67].

Another issue with wearable information is data privacy. Healthcare data from different people with diverse monitoring patterns are difficult to aggregate together to generate robust results. Patient confidentiality and data security are major concerns when using wearable devices, as ensuring compliance with HIPAA regulations can be challenging. The communication security of the collected data in Wireless Body Area Networks (WBAN) is another significant concern. Encryption is a key element of comprehensive data-centric security. Encrypted data and the use of encryption as an authentication mechanism within an organization's network are generally trusted. However, direct access to keys and certificates allows anyone to gain elevated privileges. Key management is vital to security strength, and the dependability of cryptographic schemes for key management has become an important aspect of this security. Nevertheless, the extremely constrained nature of biosensors makes designing key management schemes a challenging task.

The use of Federated Learning (FL) could mitigate several of these privacy challenges. Chen et al. extend FedHealth to develop FedHealth 2, which creates personalized models for each client by obtaining client similarities using a pre-trained model and then averaging weighted client models while preserving local batch normalization. FedHealth 2 showed increased performance in activity recognition compared to other federated methods[68]. The Federated Multi-task Attention (FedMAT) framework, built on multimodal wearable data, outperforms baseline methods in human activity recognition and is rapidly adaptable to new individuals. The framework uses an attention module at each client to learn both client-specific features and globally correlated features while ensuring data security[69]. Reddy et al. propose a blockchain-based FL system using multimodal wearable data to predict COVID-19, enabling relatively secure transmission of pertinent model development information[60].

Given the multimodal and secure nature of wearable sensor data, there is a direct need for the application of PL for building algorithms that use this data. Through the use of PL, algorithms developed using wearable data can be employed for proper remote monitoring, thereby improving patient care.

**Data Sources**
Multimodal learning has advanced through the development and curation of harmonized large-scale public datasets in the last two decades, focusing on various healthcare areas. While patchwork learning aims to utilize data privately siloed at separate organizations, methods of collection and standardization practiced by these public data consortia can be adapted when cleaning data for patchwork learning settings. For example, the Cancer Imaging Archive (TCIA), developed by the United States National Cancer Institute (NCI), houses medical images of multiple modalities (PET, CT, MRI, etc.) for over 37,000 patients, along with supplemental clinical and genomic information. Data for different tumor types and cohorts are separated via collections, many of which are publicly available and are updated continually[70]. The NCI has also curated matched tumor-normal omics data spanning 33 cancer types within the Cancer Genome Atlas (TCGA). Fueled by the data within the TCGA, multiple studies have shown innovative findings that could help develop personalized therapeutics for cancer patients. With significant overlap in sample populations across the TCIA and TCGA, both imaging, clinical, and omics modalities can be integrated to provide increased information[71]. In the field of neuroscience, the Laboratory of Neuro Imaging (LONI) has developed the Image & Data Archive (IDA), which provides resources and tools to interrogate a variety of imaging modalities, including MRI, PET, CT, and their matched clinical and genomic information[72]. The LONI IDA contains data from multiple studies conducted across the world, like the Alzheimer's Disease Neuroimaging Initiative, Human Connectome Project, and others[73,74]. While many of the featured studies have their data locked behind restricted access, LONI IDA provides methods to compare and analyze data across studies. For more general healthcare use, The Medical Information Mart for Intensive Care (MIMIC) is a large de-identified database with clinical information (vital signs, medications, survival data), billing data, and medical imaging reports of patients admitted to Beth Israel Deaconess Medical Center in Boston, Massachusetts. MIMIC-III, the most recent release of the database, builds on data collected previously, with patient data spanning over a decade. It contains 53,423 distinct adult hospital admissions across 38,597 unique patients, and several projects have been implemented to transform the MIMIC data to be processable by common data models and tools[75]. Other public and restricted-access public multimodal datasets are outlined in Table 1.

There are also various multi-institutional datasets available that make use of distributed architectures for downstream analysis. The Patient-Centered Outcomes Research Institute (PCORI) launched PCORnet in 2014, integrating data from 11 clinical data research networks, including the INSIGHT Network managed by Weill Cornell, and combining data from New York healthcare systems and the Scalable Collaborative Infrastructure for a Learning Healthcare System (SCILHS), among others. PCORnet clinical data consists of longitudinal EHR data, demographic variables, insurance claims, and biospecimen details. PCORnet research networks can facilitate the development of multi-center clinical trials, cross-sectional analysis, and retrospective data analysis[76]. Similarly, the Massachusetts Institute of Technology and Philips Healthcare leveraged multi-site data from over 200,000 critically ill patient stays in 2014-2015 to create the eICU Collaborative Research Database. This database collects data from multiple critical care units across the United States and primarily contains clinical (medication, lab tests, vital signs) and demographic information[77]. Comparable efforts have been conducted worldwide. The Scottish Safe Havens is a secure platform for researchers to perform analysis on EHR, image, and omics data across multiple silos[78]. With data governance approvals, scientists can make use of de-identified data using the Safe Haven platform. Over the years, multiple methods have been implemented to better standardize and secure data accessible through the Safe Haven, leading to over 1,000 studies. Other Safe Havens with similar infrastructure are available internationally[79,80]. Platforms like these, which encourage secure multi-site collaboration, allow for the development of robust patchwork learning formulations.

| Study | Country | Year Started | Data Modalities | Access | Sample Size (Number of Sites) |
|---|---|---|---|---|---|
| UK Biobank | UK | 2006 | Questionnaires EHR/clinical Laboratory Genome-wide genotyping WES WGS Imaging Metabolites | Open access | ~500,000 (N/A) |
| China Kadoorie Biobank | China | 2004 | Questionnaires Physical measurements Biosamples Genome-wide genotyping | Restricted access | ~500,000 (N/A) |
| Biobank Japan | Japan | 2003 | Questionnaires Clinical Laboratory Genome-wide genotyping | Restricted access | ~200,000 (N/A) |
| Million Veteran Program | USA | 2011 | EHR/clinical Laboratory Genome wide | Restricted access | ~840,000 (N/A) |
| TOPMed | USA | 2014 | Clinical WGS | Open access | ~180,000 (N/A) |
| All of Us Research Program | USA | 2017 | Questionnaires SDH EHR/clinical Laboratory Genome wide Wearables | Open access and Restricted access | 1 million (N/A) |

| Name | Country | Year | Data Types | Access | Participants (Diseases) |
|---|---|---|---|---|---|
| Project Baseline Health Study | USA | 2015 | Questionnaires EHR/clinical Laboratory Wearables | Restricted access | 10,000 (4) |
| American Gut Project | USA | 2012 | Clinical Diet Microbiome | Open access | ~25,000 (N/A) |
| MIMIC | USA | 2008 - 2019 | Clinical/EHR Images | Open access | ~380,000 (N/A) |
| MIPACT | USA | 2018 - 2019 | Wearables clinical/EHR physiological laboratory | Restricted access | ~6,000 (N/A) |
| TCIA Collections | USA | 2011 - Present | Image Clinical Genomics | Restricted access | ~37,000 (100+) |
| LONI Imaging Data Archive | USA | | Imaging (MRI, fMRI, MRA, DTI, PET) | Restricted access | ~1000 (4) |
| OASIS Brains Project | USA | 2007 - 2020 | Imaging (MR, CT, PET) Clinical Biomarker | Open access | ~5000 (4) |
| Breast Cancer Digital Repository | USA | 2010 - 2013 | Clinical Imaging | Restricted access | ~1000 |
| UTA: Medical Imaging DICOM Files Dataset | USA | 2020 | Clinical Imaging | Open access | ~1000+ (10+) |
| Alzheimer's Disease Neuroimaging Initiative (ADNI) | USA | 2017 | Imaging Clinical Genomic Biomarker | Restricted access | ~1500 (50+) |
| Australian Imaging, Biomarkers and Lifestyle (AIBL) | Australia | 2006 - 2011 | Imaging (MRI, PET) | Restricted access | ~1000 (2) |
| The Cancer Genome Atlas (TCGA) | USA | 2006 - Present | Genomic | Open access | ~20,000 (N/A) |

**Table 1.** Public or restricted access datasets useful for developing PL models

## Challenges

Patchwork learning is a potent and novel formulation for machine learning that holds the potential to revolutionize healthcare by enabling the integration of data from multiple sources and modalities. However, implementing patchwork learning systems comes with significant challenges. In this section, we will discuss some of the key challenges associated with patchwork learning and explore potential solutions and future directions for addressing these challenges. These challenges, although some are specific to either FL or MML, become even more complex to tackle when performing PL.

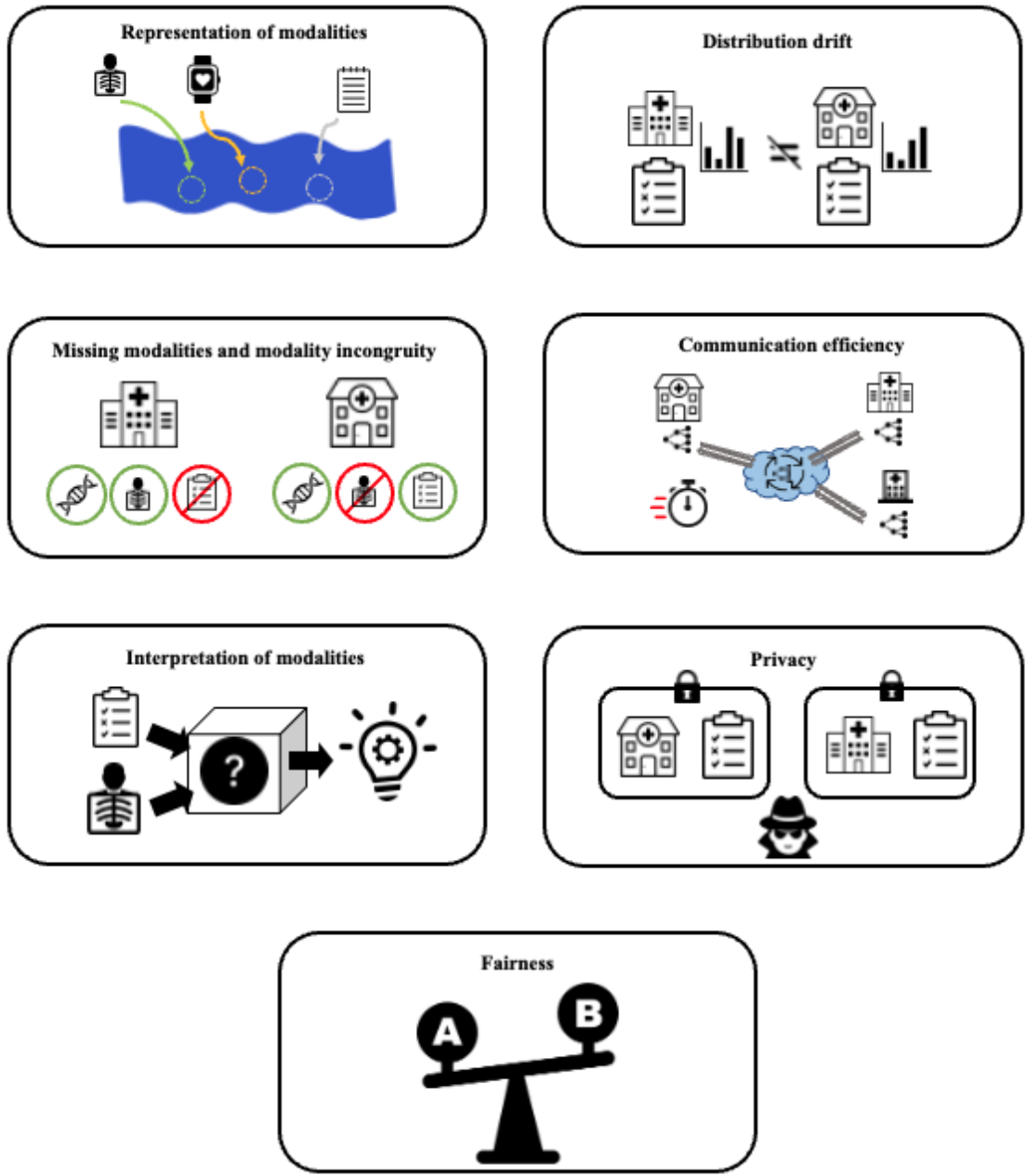

*Figure 3*: *Challenges associated with patchwork learning*

**Representation of Modalities: Effective representation of modalities are necessary for data integration and downstream processing**

With the diversity of healthcare data modalities, integrating the information concisely to allow for optimal model development is essential. Multimodal representation is the task of representing data from multiple modalities in the form of a vector or tensor. Since data from multiple modalities often contain both complementary and redundant information, the aim is to represent them in an efficient and meaningful way. Challenges associated with multimodal representation include dealing with different noise levels, missing data, and combining data from different modalities. Multimodal representation can be divided into two categories: joint and coordinated. In joint representations, the data is projected from multiple modalities into a common space, often through fusion, and in coordinated representations, each modality is projected into a separate space but they are coordinated through a similarity[81]. Currently, many MML architectures utilize fusion as a way of joining information from multiple modalities[20,82]. Recent developments like OpenAI's Contrastive Language-Image Pre-training (CLIP) demonstrate that integrating multiple modalities is essential for achieving optimal performance in machine learning tasks. CLIP is a powerful approach that uses a joint representation learning framework to learn a wide array of visual models. The CLIP model is pre-trained on a large-scale dataset of images and their associated natural language descriptions, which allows it to learn a rich and robust multimodal representation that captures both visual and textual information[83]. Multimodal large language models (LLM) also present novel ways to represent various modalities jointly. PALM-E, a multimodal model developed by Google Research, encodes real-world data into a language embedding space, allowing it to integrate various types of information from sensors. PALM-E has been used to enable effective robot training and build state-of-the-art generalized visual-language models[84]. However, while showing promise in non-specific domains, MML fusion models still face challenges in healthcare, such as being prone to overfitting[85]. To address this issue, HuggingGPT was developed, which takes advantage of the hundreds of specialized models publicly available on the Hugging Face Hub. HuggingGPT uses an LLM as a central manager that distributes sub-tasks to several downstream multimodal models to complete an overall objective. While rudimentary, these objectives can easily scale in complexity as long as certain specialized models exist downstream that are able to perform the sub-tasks[86].

Another issue is that, more often than not, modalities are not always temporally aligned and exhibit different types and levels of noise at varying points in time. These challenges arise due to the fact that different modalities are often not synchronized in time. For example, a spoken sentence may be accompanied by a facial expression which has a different temporal pattern than the speech. To address these issues, researchers have developed various approaches for temporal alignment. These approaches often involve using signal processing techniques such as time-frequency representations and attention mechanisms to detect and correct temporal misalignments[87]. Additionally, techniques such as dynamic time warping can be used to adjust timescales and achieve better temporal alignment[88]. Effective fusion of multiple modalities is also challenging due to the heterogeneous nature of multimodal data, which is only magnified when performing MML in a PL environment. To mitigate this heterogeneity, Qian et al. suggest using deep autoencoder-based nonlinear dynamical systems to learn shared information between modalities[89]. Liu et al. propose AimNet, which is the central server within a federated learning framework that performs integration and alignment between multiple modalities. The alignment and integration modules use mutual and self-attention, respectively, to generate representations that are subsequently passed on to domain-specific tasks. AimNet, however, is not shown to work when modalities are missing and also breaks privacy guarantees of FL -- representations of local data need to be sent to AimNet (central server) where a malicious party could recover raw data using those representations[90].

In a PL paradigm, co-learning methods might be preferable due to their ability to function relatively well in scenarios where modalities may be missing during training or inference[91–93]. Co-learning uses knowledge transfer from one modality to learn about a less-informed modality. Co-learning methods include the utilization of multimodal embeddings, transfer learning, multitask learning, and generative networks, with each method aiding in mitigating real-world issues with multimodal data, such as missing modalities, noisy labels, and domain adaptation[94]. While co-learning representations are not commonly used in PL settings, the advantages of co-learning over joint representations may warrant further research and development.

**Missing Modalities and Modality Incongruity: Patchwork Learning needs to account for both random and non-random missing data modalities**

The patched nature of PL arises from the real-world availability of data modalities across different healthcare sites. A common assumption in many multimodal learning models is the completeness of modalities, i.e., full and paired modalities during both training and inference. However, such an assumption may not always hold in the real world due to privacy or budget limitations. In fact, missingness is often non-random in healthcare datasets, with certain sites lacking specific data modalities due to infrastructure. Furthermore, different clinical workflows for acquiring data can also lead to non-random missingness across sites. Many investigations have explored novel methods to tackle missingness during inference time[82,92]. While less so, there have also been algorithms proposed for combatting missingness in modalities during training. Recently, generative networks (e.g., variational autoencoders, generative adversarial networks, LSTMs) have been utilized to create missing modalities. To generate one modality from another, these networks learn a joint distribution of multimodal data, through which they capture semantic correlations between modalities[95,96]. While some of these generative solutions are unsupervised, studies have shown the advantages of incorporating ground-truth labels via classification loss, certain forms of adversarial loss, or triplet loss in order to better generate modalities[97,98]. Lee et al. compared the performances of an LSTM and autoencoder architecture for generating audio modality from RGB images. Their investigation found that the incorporation of classifier loss enhanced the results of the autoencoder-based approach[99]. Confederated learning uses centrally-trained generative networks to combat missing modalities at local sites[27]. With the recent interest in diffusion models, there may be some opportunity for those architectures to synthesize missing modalities. Diffusion models are a class of machine learning algorithms that can be used to analyze how information spreads or diffuses through a network[100]. These models are commonly used to study the spread of diseases, ideas, or behaviors through social networks, but they can also be applied to other domains where information spreads through a network. One potential opportunity of using diffusion models for synthesizing missing modalities is that they can incorporate the structure of the network into the synthesis process, which can provide additional context and potentially improve the quality of the synthetic data[101–103]. However, one challenge is that diffusion models may require the availability of a network structure, which may not always be available or may need to be constructed from other sources of data. Additionally, diffusion models can be computationally intensive, especially for large networks, and may require specialized algorithms and techniques to scale to these networks. Others have explored the use of meta-learning to generate missing modalities. Ma et al. introduced SMIL, which leverages Bayesian meta-learning to perturb the latent feature space so that the embeddings of a single modality can approximate ones of full modality. Notably, SMIL utilizes significantly less data to mitigate missing modality issues as compared to solutions that use generative networks[104]. In a related field, multimodal translation has gained some attention in recent years as a potential method of both learning important semantic information from data modalities while simultaneously generating a potential missing modality[94].

A larger parallel issue that affects PL is modality incongruity, where sites may have heterogeneous data modalities available, and their local data consists of different combinations of modalities. For example, Hospital A has omics, EHR, and CT data, whereas Hospital B has omics, MRI, and wearable sensor data. In this scenario, Hospital A and B lack two modalities that the other has, but knowledge can still be derived from the complementary information of these missing modalities. This warrants the need for proper ways of learning personalizable information across multiple sites in the presence of modality incongruity. Zhao et al. propose an MMFL system where clients can have unlabeled data of different modalities, and each client trains a deep canonical correlated autoencoder to model hidden representations between modalities. The local models are aggregated in the central server through multimodal federated averaging, where a supervised model is trained using the aggregated model's encodings on an auxiliary dataset. While dealing with modality incongruity and the wealth of unlabelled data available at local clients, Zhao's framework does not take advantage of labeled data available at clients. Moreover, it requires an auxiliary dataset to be available at the central server, which is usually unrealistic in healthcare[105]. The FedMSplit framework, based on federated multi-task learning (FMTL), trains over multimodal distributed data without assuming similar modalities in all clients. FedMSplit employs a dynamic and multi-view graph structure to adaptively capture the correlations amongst multimodal client models. Client models are split into smaller shareable blocks where each block provides a specific view of client relationships. The multi-view graph captures and shares correlations between clients as edge features, through which personalized, but globally-correlated, multimodal client models can be learned. While model-architecture agnostic and adept at handling non-IID data, FedMSplit is not able to make use of unlabeled data at sites. This limits the framework's applicability in healthcare tasks where labeled data is sparse[106].

A critical concern when addressing modality incongruity in patchwork learning is the preservation of data connectivity. To align the semantic feature space between different data modalities, it is essential that no block on the patchwork is isolated, meaning a particular modality must not be present exclusively at one site. Isolated blocks may hinder the identification of complementary information between modalities, thus impacting the efficacy of the learning framework. While most current investigations do not face significant data connectivity issues due to the limited number of modalities involved, real-world systems that incorporate 10 or more modalities are more susceptible to such challenges. For instance, Site A may possess accelerometer data from a wearable sensor, which is unavailable at other sites. This isolation restricts the ability to determine how the accelerometer data could complement other modalities. Administratively, this could necessitate a decision on whether to include Site A in the patchwork learning framework. In scenarios where a PL setting has isolated blocks of data, external sources of connective information will be needed to properly perform training. One solution for connecting these isolated blocks is through multimodal generation via LLMs. LLMs excel at transfer learning and domain adaptation, which enables them to transfer knowledge from one domain or task to another with minimal labeled data. This capability can be leveraged to establish connections between isolated data blocks and adapt models to site-specific tasks. The general-purpose representations learned by LLMs can be fine-tuned on specific medical tasks or modalities, adapting the models to the unique requirements and nuances of healthcare applications. This process of fine-tuning can help LLMs learn to better connect isolated data blocks and facilitate the extraction of complementary information across modalities. Recent advancements in this field include the development of Generative Pre-trained Transformer 4 (GPT-4), HuggingGPT, and PalmMED, among others[86,107,108]. These models demonstrate the potential for LLMs to support patchwork learning frameworks in addressing modality incongruity and data connectivity challenges. Necessary connective data can also be extracted from large-scale knowledge graphs like the Integrative Biomedical Knowledge Hub and the Clinical Knowledge Graph[109,110]. Biomedical knowledge graphs can be a valuable resource for connecting disparate datasets in patchwork learning. These graphs are large-scale, structured networks of biomedical information that can be used to represent and link various concepts, entities, and relationships in the domain of health and medicine. By using knowledge graphs, researchers can extract external information that can be used to connect different datasets in a patchwork learning setting. For example, knowledge graphs can be used to identify shared concepts or entities between different datasets, such as specific diseases, drugs, or genes. This information can be used to map the data from different datasets onto a common ontology or feature space, allowing the data to be more easily combined and used for training machine learning models. Additionally, knowledge graphs can be used to provide contextual information about the data, such as the relationships between different entities or the attributes of specific concepts. This can help improve the accuracy and interpretability of the machine learning models and can also support the development of more complex and sophisticated models that can better capture the complex relationships and dynamics of health and disease. The use of external information extracted from biomedical knowledge graphs can be a valuable approach for connecting disparate datasets in patchwork learning. To do so, knowledge graphs need to be multimodal and need to be able to adeptly link information across modalities[111,112]. By providing a rich and structured representation of the biomedical domain, knowledge graphs can support the integration and analysis of data from multiple sources and can help enable the development of more accurate and effective machine learning models for healthcare applications. Steps have already been taken in generating multimodal biomedical knowledge graphs that provide valuable sources of external information for downstream model development. Zhao et al., using numerous data sources, created a multimodal knowledge graph with a host of suites that enables users to pose questions to the graph and retrieve appropriate answers. Questions can be posed via multiple modalities, enabling varied usage of the framework. Similarly, LingYi is another medical knowledge graph that does the same, allowing multimodal translation through some of its modules[113]. Future iterations of systems like these may enable PL frameworks that are missing modalities to generate bridges by querying knowledge graphs[114].

**Interpretation of Models: Patchwork learning formulations need to be interpretable for healthcare application**

While seeking high performance is generally a reason to warrant the use of PL, this can be at the cost of interpretability of the subsequent model. There is significant interest in understanding the complex cross-modal associations in diagnostic decisions to further uncover hidden disease mechanisms, facilitate understanding of the disease, and build trust in statistical models. In clinical decision-making, interpretability of models is especially important, as several checks and balances need to be established when generating diagnoses or providing recommendations. Interpretability should seek to address both modality-specific contributions and inter-modality interaction contributions[115]. Simple approaches that have seen some success involve treating each modality separately when determining the post-hoc interpretability of the modality. Han et al. train a multimodal network for estimating postoperative pain and use SHapley Additive exPlanations (SHAP) on the fused multimodal feature space to obtain model explanations. While providing some general information about modality contribution, it fails to provide relevant information about each modality that could be clinically evaluated[116]. Moreover, methods like these are limited in their ability to explain the contribution of complementary information instrumental in multimodal model performance. Others have developed modality-agnostic methods through post-hoc model interpretation. DIME (Fine-grained Interpretations of Multimodal Models via Disentangled Local Explanations) provides explanations for model predictions by disentangling the contributions of a model into those that are due to unimodal contributions and multimodal interactions. By doing so, clinicians can identify what facets of the overall model a prediction is based on. Although DIME is model-agnostic, it has only been shown to work on models that provide discrete outputs. Moreover, as the number and diversity of modalities increase, the cost of disentanglement and interaction explanation becomes exponentially larger[117].

Other multimodal networks are intrinsically interpretable through model design. These include graph-based fusion techniques, multimodal explanation networks, neuro-symbolic reasoning, or attention-based methods[93,118,119]. These approaches individually focus on building interpretable components for either modality or modality interaction. Attention-based approaches, where weights are assigned to different input features, have attracted significant attention recently. However, the explanatory power of these mechanisms is questionable since there is often a lack of association between gradient mappings and attention weights[120–122]. In general, these methods suffer from only working due to careful model design and are limited to providing explanations only on specific modalities. However, with the distributed nature of PL, several of these methods may be limited in providing accurate interpretations. Lack of access to cross-client data limits several interpretation mechanisms in their ability to provide both global and local explanations.

Causality is a crucial aspect in enhancing the interpretability of models, as causal relationships are inherently comprehensible to humans. Causal machine learning facilitates the investigation of a system's response to an intervention (e.g., outcomes given a treatment in the healthcare domain). Quantifying the effects of interventions (causal effects) enables the formulation of actionable decisions while maintaining robustness in the presence of confounders[123]. In the context of PL, multi-modal data can serve as proxies for unobserved confounders, thereby improving the accuracy of causal effect estimation[124]. Addressing the missingness of modalities is a vital consideration for this objective. Furthermore, estimating the heterogeneous causal effects across different sites presents a challenge for causal machine learning within the PL framework[125].

As such, future research efforts in the realm of interpretability should aim to develop methods that are more widely applicable, efficient, and effective at providing detailed explanations for PL-based models. This may involve the exploration of novel model architectures or interpretation techniques that are capable of capturing and disentangling the contributions of multiple modalities in a more comprehensive manner. Additionally, research should focus on enhancing the scalability and efficiency of interpretation methods in the context of distributed learning, enabling their application to larger and more complex and distributed multimodal datasets. In summary, the continued development and improvement of interpretability methods for PL-based models will be crucial for the advancement of these models in clinical decision-making and other applications.

**Distribution Drift: Heterogeneity between sites needs to be accounted for in patchwork learning**
Considering the regional disparities among participating healthcare facilities, the distribution of data across clients can vary significantly. This not only results in sample heterogeneity and non-IID data dispersed across sites, but also leads to potential variations in the relationships between input features from one site to another. In PL, the existence of multiple modalities and the potential absence of some modalities at specific locations further exacerbate the challenge of addressing distribution drift. For instance, PET scans may be captured using distinct scanners and protocols at various sites, leading to differences in image resolution, size, and inter-slice spacing. Consequently, the relationships between these PET scans and their corresponding site-specific electronic health record (EHR) data may differ. Distribution drift has been identified as a primary factor contributing to model performance degradation and unfairness in distributed architectures and PL frameworks, necessitating additional communication rounds for multi-modal federated learning (MMFL) systems to achieve convergence[126].

Domain generalization is one machine learning area that addresses distribution drift. Specifically, domain generalization presumes the existence of data from multiple source sites. Several methods have been proposed for training a model utilizing multi-source data, ensuring generalizability to any unseen site[127]. Muandet et al. suggested learning an invariant transformation of the input by minimizing dissimilarity across domains while preserving the functional relationship between input and output variables[128]. Furthermore, the authors provided a learning-theoretic analysis demonstrating that reducing dissimilarity enhances expected generalization in new domains. This objective can also be accomplished through adversarial training[129]. Additionally, some studies have focused on learning an invariant transformation of the conditional distribution of the input given the outcome class, rather than the input itself[130]. These approaches prove effective in addressing conditional shifts across sites.

In addition to the aforementioned methodologies, a series of techniques known as domain invariant learning have been proposed to address domain generalization. Invariant risk minimization (IRM), introduced by Arjovsky et al., aims to reduce the effect of spurious, or non-causal, properties within different sets of training data. In a setting where training data is split into multiple separate environments with their own site-specific biases, invariant risk minimization promotes the learning of features that are stable across sites[131]. More specifically, IRM starts by defining a set of tasks that the model needs to learn. Each task is associated with a different distribution of input data, and the goal is to learn a model that performs well on all of these tasks. The model is trained by minimizing a loss function that combines a ML model's standard loss with a penalty term that encourages invariance across the different tasks. This penalty term is designed to measure the difference between the model's predictions on two different tasks, and it is minimized when the model produces similar outputs for similar inputs, regardless of the task. Zare et al. introduced ReConfirm, which extends the IRM framework by accounting for class conditional variants and shows significant improvements over traditional trained ML models on medical data[132]. In the context of PL, IRM can be used to train a model that integrates information from multiple datasets that are distributed across separate sites and contain different modalities. Specifically, IRM can be used to learn a set of features that are consistent across different datasets, even if they contain different modalities or have different patient populations. By doing so, IRM can promote the generalization of the model to new data modalities and patient populations, thereby reducing distribution drift across sites. To implement IRM in patchwork learning, the datasets from different sites can be treated as different tasks, and the model can be trained to perform well on all of these tasks by minimizing a loss function that includes a regularization term that encourages invariance across different datasets. The regularization term can be based on a measure of the similarity between the distributions of the model's output across different datasets, such as the maximum mean discrepancy (MMD) measure. That said, an IRM formulation does pose challenges in settings where the relationship between inputs and outputs is non-linear, providing no significant improvement over standard training[133]. Further adaptations of IRM will be necessary to adapt it to real-world settings that use complex multi-site healthcare data.

In recent years, there has been a push to generate personalized, globally correlated models to mitigate drift across clients and data modalities. Personalized federated models are grouped into two categories: global model personalization and local-level personalization[14]. Global model personalization trains a single global model which is subsequently personalized for each client through local adaptation. One implementation of global model personalization is Per-FedAvg and its extension pFedMe[15,134]. Per-FedAvg uses model-agnostic meta-learning (MAML) to formulate FedAvg into developing an initial global model that performs well on heterogeneous clients with only a few steps of gradient descent. Chen et al. propose hierarchical gradient blending (HGB), which adaptively calculates an optimal blending of modalities to minimize overfitting and promote generalization. HGB is task and architecture agnostic and shows promise in mitigating lack of generalization in MMFL. However, initial implementations of HGB are not able to make the most use of complementary information between modalities[135].

Local-level personalization can be further divided into two categories: architecture-based and similarity-based approaches. Architecture-based approaches enable personalization by designing different models for each client, whereas similarity-based approaches seek to identify client relationships and provide related clients with similar models[14]. FedMD, an architecture-based approach, allows for the creation of personalized, architecture-agnostic models at clients through the use of transfer learning and knowledge distillation. While the architecture-agnostic aspect of FedMD potentially allows the incorporation of differing modalities at different clients, FedMD requires a public dataset, which is infeasible in many healthcare scenarios[136]. Lu et al. use FedAP to mitigate heterogeneity across clients by calculating the similarity between clients based on batch normalization weights. FedAP creates personalized models with less communication cost and has been evaluated on several healthcare datasets[137]. An extension of these classes of architectures, FedNorm utilizes the minibatch normalization (MN) technique, an extension of using batch normalization, to create personalized models in the presence of data heterogeneity and to combat modality incongruity. The framework normalizes feature information by modality before distribution across all clients. FedNorm allows clients to have a mix of modalities while simultaneously combating data heterogeneity by building personalized models. However, FedNorm has shown success only when the modalities available are all of the same data type (e.g., PET scan and MRI -- both images)[138]. A popular similarity-based approach for local-level personalization is through federated multi-task learning, where a model jointly performs several related tasks by leveraging domain-specific knowledge across clients. FMTL has shown promise in building models in federated settings with the MOCHA and FedAMP algorithms and more recently, PL-based models through techniques like FedMSplit[135,139,140]. In a recent study, Collins et al. introduce FedRep, a novel federated learning framework and algorithm, for the purpose of learning shared representations across distributed clients and unique local heads for each client. FedRep addresses the challenge of biases in current machine learning models by incorporating data from different modalities and sources, resulting in a shared feature representation that can be applied to a variety of tasks. FedRep's ability to learn shared low-dimensional representations among data distributions makes it useful for meta-learning and multi-task learning in patchwork learning settings[141].

Due to the fast-paced nature of healthcare data, PL-based architectures need to be aware of distribution drift that occurs over time, also known as concept drift[142,143]. For example, a PL-based model developed in 2022 may no longer be accurate for a population from the same hospital in 2023. Methods to mitigate or adapt PL-based models to time-dependent distribution drifts are necessary for the long-term use of these models in the real world. Currently, solutions to mitigate concept drift have utilized drift detection followed by time-varying clustering or other adaptation mechanisms[126,144,145]. Another popular formulation for addressing distribution drift due to time is continual learning, also known as lifelong learning or online machine learning. Continual learning enables models to continuously learn and evolve based on increasing amounts of data while retaining previously learned knowledge. This allows them to incrementally learn and autonomously change their behavior without forgetting the original task[35,146]. However, continual learning faces several obstacles in practice, particularly in healthcare settings where the stakes for real-time medical applications of AI are high due to their impact on health outcomes. One of the primary obstacles is catastrophic forgetting, where new information interferes with what the model has already learned, leading to a decrease in performance or an overwrite of the model's previous knowledge. To address these challenges, online training methods that do not involve full retraining but rather the use of new data only, are likely to be more realistic in the healthcare setting. However, implementing continual learning models in the clinical arena requires consideration of other relevant challenges, including the absence of established methods for assessing the quality of these models. Validation of these models needs to encompass factors such as the collection process for new data, the automated organization or labeling of new data, knowledge transfer between new and original data, and the overall performance of the model after incorporating data, while ensuring that no catastrophic interference occurs.

**Communication Efficiency: Techniques to minimize lag due to communication are instrumental for patchwork learning infrastructure**

Communication is a key bottleneck to consider when developing methods for federated networks, especially in PL systems. This is because PL systems could include a massive number of sites or individual patient silos, and communication in the system can be slower than local computation by many orders of magnitude. With the integration of multiple data modalities, computation time increases as many current multimodal learning (MML) techniques require significant amounts of preprocessing and/or communication to integrate. Therefore, real-world PL-based models will have to rely on communication-efficient methods. Generally, there are two broad principles for improving communication efficiency: reducing the total number of communication rounds or reducing the size of transmitted messages at each round.

One class of methods focuses on optimizing the local updating processes, allowing for a variable number of local updates to be applied on each machine in parallel at each communication round. The goal of local updating methods is to reduce the total number of communication rounds. Guha et al. introduce one-shot federated learning (FL), where the central server only requires a single round of communication to learn a global model through the use of ensembling and model distillation[147]. Zhou et al. expand one-shot FL through DOSFL, in which each client distills their data to be sent to the central server, where a global model is trained[148]. COMMUTE utilizes transfer learning and distance-based adaptive regularization to create a one-shot multi-site risk prediction framework. While the method mitigates the effects of data heterogeneity across sites, it limits all clients to using the same set of features. Moreover, its performance with complex and deep model architectures remains unknown[149].

Another class of methods that has seen success in decreasing communication costs is model compression, which includes sparsification, subsampling, and quantization. Zhang et al. introduce dynamic fusion-based federated learning to choose participating clients according to local performance, thereby improving communication efficiency. They applied this method to predict COVID-19 across secure nodes and showed performances comparable and/or higher than FedAvg on different facets[150]. Recently, decentralized training has garnered a lot of attention for its ability to increase communication efficiency. While standard FL settings require a central server for connecting all remote devices and performing aggregations, decentralized FL systems provide an alternative when communication to the server becomes a bottleneck, especially when operating in low bandwidth or high latency networks[4].

**Privacy: Patchwork learning should have infrastructure and methods to ensure client privacy**

Data privacy is of utmost importance in healthcare, particularly when it comes to training machine learning models. Patient data is highly sensitive and must be protected to maintain trust and confidentiality. However, machine learning models require large amounts of data to be effective, creating a tension between privacy and innovation. Therefore, developing privacy-preserving machine learning techniques for healthcare can help mitigate these concerns and enable the development of accurate models while preserving patient privacy[7,8]. While patchwork learning may provide some data confidentiality by limiting data sharing, adversarial attacks have revealed possible vulnerabilities. For instance, Carlini et al. have shown that unwitting memorization of neural networks from the training dataset may reveal personally identifiable information[151]. Moreover, models themselves may contain intellectual property (IP), and the learned parameters of the models can reveal valuable information about the model's architecture, design, and functionality, which could be used by others to replicate or reverse-engineer the model without the owner's permission. If the information within the datasets is leaked through machine learning models, it could not only harm privacy but also undermine trust in such collaborative implementations[152].

There are a myriad of third-party attacks that can be used against federated ML architectures to exploit the parameters and correspondingly access the training data. These include membership inference, model inversion, property inference, and functionality stealing attacks. Membership inference attacks are conducted by a malicious actor with limited access to a model and entail using the characteristics of the model to determine if a given data point is included in the training dataset. In this attack, a Generative Adversarial Network (GAN) can be trained on samples generated from a target model and correspondingly learn from the characteristics and statistical distributions associated with the original training data, making it capable of recreating the training data. Hitaj et al. have shown the efficacy of this approach[153]. Model inversion techniques aim to infer class features given the limited access the adversary has to a model's gradients and are successful at extracting information in certain settings. Property inference attacks look to infer properties about an entire class of training data. These properties are not intentionally leaked by the models; rather, they are just artifacts of the learning process. Lastly, in functionality stealing attacks, adversaries aim to extract model parameters through exploitation of the model's outputs[154,155]. These adversarial attacks and their derivatives have fueled the increasingly dire issue of model and data protection.

In healthcare, privacy-preserving methods are essential for ensuring the privacy of customer information, as data used to train machine learning models could be compromised and exploited through an attack from adversaries. With the introduction of federated learning methods to mitigate issues with data heterogeneity and data sharing challenges, these aforementioned attacks have become more viable. Secure Multi-party computation (SMPC), differential privacy, and homomorphic encryption were introduced to prevent malicious attacks[152]. Each of these methods has its advantages and shortcomings. SMPC and homomorphic encryption are computationally costly and require complex infrastructure to maintain[156]. With differential privacy, the performance of machine learning models is usually compromised for increased security[157]. Moreover, while differential privacy and other obfuscation techniques have shown some promise in unimodal data, there are adaptation issues when working in a multimodal learning (MML) setting. These methods are often fine-tuned for specific scenarios or model algorithms. The diversity of data modalities generates different definitions of differential privacy algorithms, which leads to difficulties in unifying them into one algorithm[158].

Recently, the large-scale use of blockchain has provided researchers with another method for increasing the security of distributed systems. Chang et al. designed a blockchain-based federated learning (FL) framework for medical IoT devices, which utilizes differential privacy and gradient-verification protocols to catch poisoning attacks. When tested on the task of diagnosing diabetes based on EHR data, their architecture is able to limit the success of poisoning attacks to less than 20%[159]. Another framework, proposed by Rehman et al., uses blockchain and an intrusion detection system (IDS) to detect malicious activity during model training within a federated healthcare network. The end-to-end system allows for models to be developed on several modalities, ranging from medical IoT data to medical images, and gives physicians the ability to monitor patient risk for diseases in real-time. While currently limited in its computational complexity, the framework shows promise in providing a system for medical organizations to develop risk prediction models based on multimodal data[160].

Swarm learning (SL) uses blockchain technology to combine decentralized hardware infrastructures to securely onboard clients and dynamically generate a global model, whose performance is comparable to models trained when all data is pooled. Through the use of blockchain technology, SL is able to mitigate the harm of dishonest participants or adversaries attempting to undermine the network. SL has been used in proof-of-concept applications to predict COVID-19, leukemia, and other pathologies in a setting where clients have non-IID data. While decentralization greatly enhances the network's resistance to attacks and data heterogeneity, SL loses capabilities other frameworks have due to the lack of central aggregators. Moreover, current implementations of SL may be affected by latency between clients, slowing calculation transportations[161].

The integration of novel privacy-preserving technologies and architectures is crucial for the successful implementation of patchwork learning systems. By implementing these techniques, PL systems can protect the sensitive data being used and help ensure that it is only used in appropriate and ethical ways. This is essential for building trust in these systems and ensuring that they can be used safely and effectively in a wide range of settings. Additionally, the use of privacy-preserving technologies and architectures can aid in following relevant privacy laws and regulations and can also support the development of new machine learning models and applications that can improve the health and wellbeing of individuals and communities.

**Fairness: Patchwork learning settings should enforce fairness across clients**
One major challenge of federated learning (FL), which becomes even more difficult in patchwork learning, is achieving collaborative fairness among participating clients. Each client's contribution to the central model is usually far from equal due to various reasons, with the primary reason being distributional discrepancies across different clients. In certain scenarios, some clients may be negatively affected through distributed learning[14]. As machine learning (ML) models are deployed in increasingly important applications, ensuring that the trained models do not discriminate against sensitive attributes has become another critical factor for federated learning. In general, fairness in PL falls under two categories: 1) performance fairness, where every client sees a performance increase from participating, and 2) collaboration fairness, where participants with higher contributions receive higher rewards or incentives. These incentives can include reputation, monetary compensations, or additional computational infrastructure, among others[162,163]. While not all categories of fairness need to be exercised in all clinical settings of PL, algorithms that further these principles should be available for real-world use.

To ensure performance fairness, Li et al. propose a q-Fair FL framework to achieve an improved uniform accuracy distribution across participants at the cost of model performance. q-Fair FL utilizes a novel optimization technique that reweights local objectives, which was inspired by resource allocation strategies in wireless networks[163]. Agnostic federated learning optimizes a model for any target distribution formed by a mixture of clients' distributions, forcing the model not to overfit to any particular client[164]. Hao et al. propose Fed-ZDAC and Fed-ZDAS, which utilize zero-shot data augmentation (generating synthetic data based only on model information rather than sample data points) on under-represented data to decrease statistical heterogeneity and encourage uniform performance across clients[165]. Other methods have utilized multitask federated learning and other personalization techniques to achieve performance fairness by mitigating the presence of data heterogeneity, which is often the root cause of non-uniform performance[14,166].

Collaborative fairness is essential when there are discrepancies in contributions between PL clients. These contributions can vary due to data volume, data quality, computation power, and potential risks that each client takes by participating in PL. A fair collaborative environment is one where each participant receives a reward that fairly reflects its contribution to the PL system. When developing collaborative fair environments, measurement of contribution, reward for contribution, and distribution of reward all need to be determined. The federated learning incentivizer (FLI) was proposed as a payoff-sharing scheme to achieve contribution and expectation fairness. FLI is formulated to work with any definition of contribution and cost but is primarily used for monetary rewards, which is not the norm in healthcare scenarios[167]. Robust and Fair Federated Learning (RFFL) distributes better performing models to clients who have higher contributions. RFFL calculates a 'reputation' for each client which designates how much contribution each client provides; clients with contributions lower than a certain threshold are removed[168]. Cui et al. introduce the idea of collaboration equilibrium, a setting where clients are clustered together, where there are no other settings that individual clients can benefit more. They utilize a Pareto optimization framework using benefit graphs to generate clusters of clients that achieve collaboration equilibrium. While showing promise in achieving collaborative fairness, the framework requires all local clients to agree on constructing a benefit graph by an impartial third-party before model training occurs[162].

In general, creating generalizable fair architectures for PL is difficult for two reasons. The necessity of privacy-preservation measures in PL limits access to sensitive variables from all parties, which is required to properly evaluate fairness. Applying fairness constraints locally on each client will lead to inferior fairness performance or null solutions due to the inaccurate model fairness measurements computed locally as a result of distributional differences among clients and the resulting conflicts between local fairness constraints. Finding a proper fairness measurement has proven to be a difficult task as well, with each measurement having its own disadvantages and advantages. With the addition of various modalities of data, each of which may be asynchronously implemented and integrated into local architectures, the difficulty of finding generalizable and fair PL-based architectures increases. In the future, approaches to fairness in patchwork learning will likely continue to be an important focus of research, as the field continues to evolve and develop new methods for integrating data from multiple sources.

## Conclusion

Patchwork learning is a powerful and innovative framework with the potential to transform healthcare by enabling the integration of data from multiple sources and modalities. Allowing researchers to combine data from disparate datasets and sites, patchwork learning can provide a more comprehensive and accurate view of health and disease, supporting the development of new and improved diagnostic, predictive, and therapeutic tools. However, significant challenges are associated with implementing patchwork learning, including issues related to communication efficiency, privacy, and fairness. To overcome these challenges and fully realize the potential of patchwork learning in healthcare, it is crucial for researchers to continue exploring and developing new methods and technologies that can improve the performance and scalability of these systems. This involves a combination of efforts to optimize local updating processes, reduce communication costs, and enhance the privacy and security of data used in patchwork learning. Additionally, continued research on fairness and bias in patchwork learning will be critical to ensuring that these systems produce accurate and equitable results. Overall, patchwork learning represents a promising and exciting new direction in machine learning and healthcare. By enabling the integration of data from multiple sources and modalities, patchwork learning has the potential to support the development of new and improved tools for diagnosing, predicting, and treating health conditions, ultimately helping improve the health and wellbeing of individuals and communities.

## Acknowledgement
This work is supported by NSF awards with number 1750326, 2212175, and NIH awards with number RF1AG072449, R01MH124740 and R01AG080991.

## Author Contributions
S.R. and F.W. conceptualized the study. S.R. and W.P. drafted the manuscript. M.R., Y.C., and J.Z. provided substantial contributions to the design of the work, critically revised the manuscript for important intellectual content. All the authors read the paper and suggested edits. F.W. supervised the project, critically revised the manuscript, and gave final approval of the version to be published.

## Competing interests
The authors declare no competing interests.